\documentclass[letterpaper, preprint, paper,10pt]{AAS}	

\usepackage{bm}
\usepackage{amsmath}
\usepackage{subfigure}
\usepackage[colorlinks=true, pdfstartview=FitV, linkcolor=black, citecolor= black, urlcolor= black]{hyperref}
\usepackage{overcite}
\usepackage{footnpag}			      	
\usepackage{caption}
\usepackage{floatrow}
\usepackage{tablefootnote}
\usepackage[T1]{fontenc}
\usepackage[utf8]{inputenc}
\usepackage{babel}

\PaperNumber{23-303}

\usepackage{acronym}

    \acrodef{APL}{Applied Physics Laboratory}
    \acrodef{GSFC}{Goddard Space Flight Center}

    \acrodef{TCI}{Titan-Centered Inertial}
    \acrodef{TCTF}{Titan-Centered Titan Fixed}
    \acrodef{NED}{North-East-Down}
    \acrodef{TOF}{Take-off Frame}
    \acrodef{FRD}{forward-right-down}
    \acrodef{MECH}{mechanical}

    \acrodef{IMU}{inertial measurement unit}
    \acrodefplural{IMU}[IMUs]{inertial measurement units}
    \acrodef{INS}{Inertial Navigation System}
    \acrodef{GPS}{Global Positioning System}
    \acrodef{GNSS}{Global Navigation Satellite System}

    \acrodef{KF}{Kalman Filter}
    \acrodef{EKF}{Extended Kalman Filter}
    \acrodef{MEKF}{Multiplicative Extended Kalman Filter}
    \acrodefplural{MEKF}[MEKFs]{Multiplicative Extended Kalman Filter}
    \acrodef{GMM}{Gaussian Mixture Model}
    \acrodef{NLLS}{Nonlinear Least Squares}
    \acrodef{SLAM}{Simultaneous Localization and Mapping}
    \acrodef{MAP}{maximum a posteriori}
    \acrodef{m-estimators}{Maximum Likelihood Estimators}

    \acrodef{MAP}{maximum a posteriori}
    \acrodef{MLE}{maximum likelihood estimate}
    \acrodef{PSD}{positive-semidefinite}
    \acrodef{FOGM}{First-Order Gauss Markov}
    \acrodef{DCM}{direction cosine matrix}
    \acrodefplural{DCM}[DCMs]{direction cosine matrix}

    \acrodef{RSOS}{Residual-Sum-of-Squares}

    \acrodef{CONOPs}{concept of operations}
    \acrodef{VIO}{visual inertial odometry}
    \acrodef{NavCam}{navigation camera}
    \acrodef{FIFO}{first-in-first-out}
    \acrodef{EDL}{entry, descent, first flight, and landing}
    \acrodef{ETS}{Electro-optical Terrain Sensing}
    \acrodef{MGNC}{Mobility Guidance, Navigation, and Control}
    \acrodef{AGL}{above ground level}

    \acrodef{DEM}{digital elevation model}

    \acrodef{CPU}{Central Processing Unit}

\begin{document}

\title{Preliminary Design of the \\ Dragonfly Navigation Filter}

\author{Ben Schilling\thanks{Johns Hopkins Applied Physics Laboratory, 11100 Johns Hopkins Road Laurel, MD 20723}, Timothy G. McGee\thanks{Point Mass Technologies, Pittsburgh, PA}, Ryan Mitch\footnotemark[1], and Ryan Watson\footnotemark[1] \\ }

\maketitle{}

\begin{abstract}
 Dragonfly is scheduled to begin exploring Titan by 2034 using a series of multi-kilometer surface flights. This paper outlines the preliminary design of the navigation filter for the Dragonfly Mobility subsystem. The software architecture and filter formulation for lidar, visual odometry, pressure sensors, and redundant IMUs are described in detail. Special discussion is given to developments to achieve multi-kilometer surface flights, including optimizing sequential image baselines, modeling correlating image processing errors, and an efficient approximation to the Simultaneous Localization and Mapping (SLAM) problem. 
\end{abstract}
\section{Introduction}

Dragonfly will investigate prebiotic chemistry and habitability on Saturn's largest moon, Titan, in the mid-2030s.\cite{barnes2021science,lorenzTitan} Successful exploration of Titan will require a series of multi-kilometer surface flights to access materials in locations with diverse geological histories.\cite{lorenz2018dragonfly, mcgee2018guidance} This paper highlights the preliminary design of the navigation (nav) filter for the Mobility subsystem, which is responsible for the guidance, navigation, and control (GNC) of the relocatable rotorcraft. In addition to operating autonomously, navigating on Titan presents several unique challenges. First, no external infrastructure or orbital assets are available for navigation. Surface maps constructed from Cassini flybys are too coarse for navigating near the surface, and the presence of diverse terrain and unknown topography require robust algorithms with limited a priori knowledge.~\cite{Lorenz_2021} Further, there are no external attitude references (e.g. stars) near the Titan surface, nor does a useful magnetic field exist, precluding the use of external sensors for attitude knowledge. 

Since high resolution maps of the Titan surface are not available, Dragonfly leverages \ac{VIO}, where an \ac{IMU} and sequential images are used to estimate how far the vehicle has traveled. \ac{VIO} has been successfully flown on NASA's MER-DIMES and Ingenuity missions.\cite{MER, ingenuityNav} One limitation of \ac{VIO} is the lack of observability of the absolute vehicle position, causing position errors to accumulate over time. On Dragonfly the impact of accumulated error is mitigated by periodically saving an image for future navigation.\cite{mcgee2018guidance}  The saved images, or "breadcrumbs", are used to navigate back to the take-off site, or to a pre-scouted landing site using the leapfrog approach.\cite{mcgee2018guidance, witte2019no, schilling2019} When revisiting breadcrumbs, the nav filter must fuse correlated position measurements, a variation of the \ac{SLAM} problem.\cite{smith1990estimating, moutarlier1991incremental} The traditional \ac{SLAM} formulation of maintaining many landmarks (i.e. breadcrumbs) in the filter state and updating landmarks during revisits is intractable on the flight processor selected for Dragonfly, serving as motivation for an efficient and robust approximation to the \ac{SLAM} formulation.

This paper focuses on design challenges related to long distance traverse navigation on Titan. The \ac{EDL} sequence offers different challenges for navigation including large initial attitude errors with respect to gravity and high angular rates throughout the parachute descent phase. Navigation for \ac{EDL} is achieved using the same nav filter described in this paper and will be the topic of a future paper. The organization of this paper follows as: the~\nameref{background} section discusses material relevant to the Dragonfly mission and nav filter design; the~\nameref{design} section outlines the preliminary design, including the architecture, state space, and measurement models; the~\nameref{results} section illustrates performance from a high fidelity Titan simulation, and finally the~\nameref{conclusion} section summarizes results and highlights future work.
\section{Background} \label{background}

The Dragonfly relocatable lander, illustrated in Figure~\ref{fig:dragonfly}, is equipped with two \acp{IMU}, two pressure sensors, one lidar, and one \ac{NavCam}. The sensor rates and measurement types are listed in Table~\ref{tab:sensor}. Further details of the baseline sensor specifications are captured in Table~\ref{tab:sim_setup} of the \nameref{results} section.

\begin{minipage}{\textwidth}
  \begin{minipage}[b]{0.49\textwidth}
    \centering
    \includegraphics[width=3in]{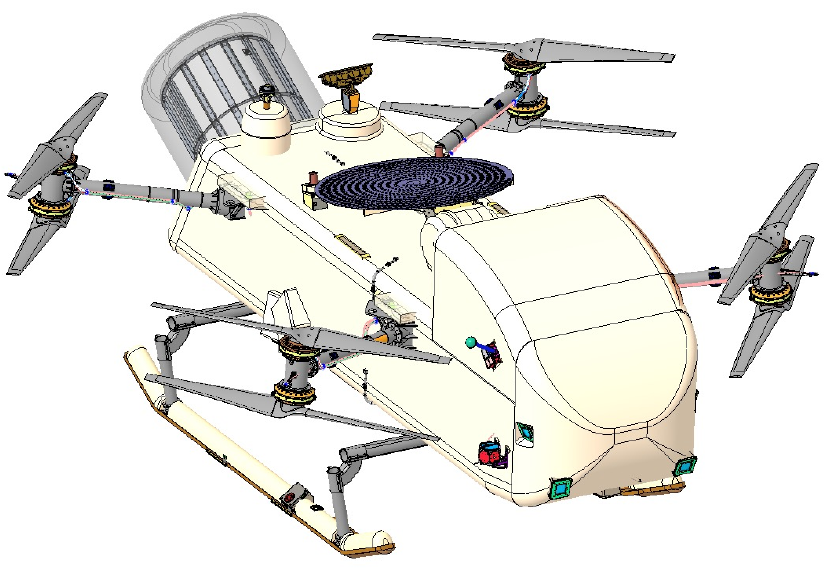}
    \captionof{figure}{Dragonfly relocatable rotorcraft}
    \label{fig:dragonfly}
  \end{minipage}
  \hfill
  \begin{minipage}[b]{0.49\textwidth}
  	\fontsize{9}{8}\selectfont
    \centering
   \begin{tabular}{| c | c | c |} 
      \hline 
      Sensor & Rate (Hz) & Measurement \\
      \hline 
     IMU (x2) & 200 & $\Delta$V,$\Delta\theta$ \\ 
      Pressure (x2) & 10 & Absolute pressure\\
     Lidar & 5-10\tablefootnote{The lidar runs at higher rates during hazard detection scans, but only a subset of data is used by the nav filter} & X,Y,Z point \\
     NavCam/ETS & 1 & Position displacement \\      \hline
   \end{tabular}
      \captionof{table}{Dragonfly sensor suite}
      \label{tab:sensor}
      \vspace*{0.6in}
    \end{minipage}
\end{minipage}

\begin{figure}[htbp]
	\centering\includegraphics[width=6in]{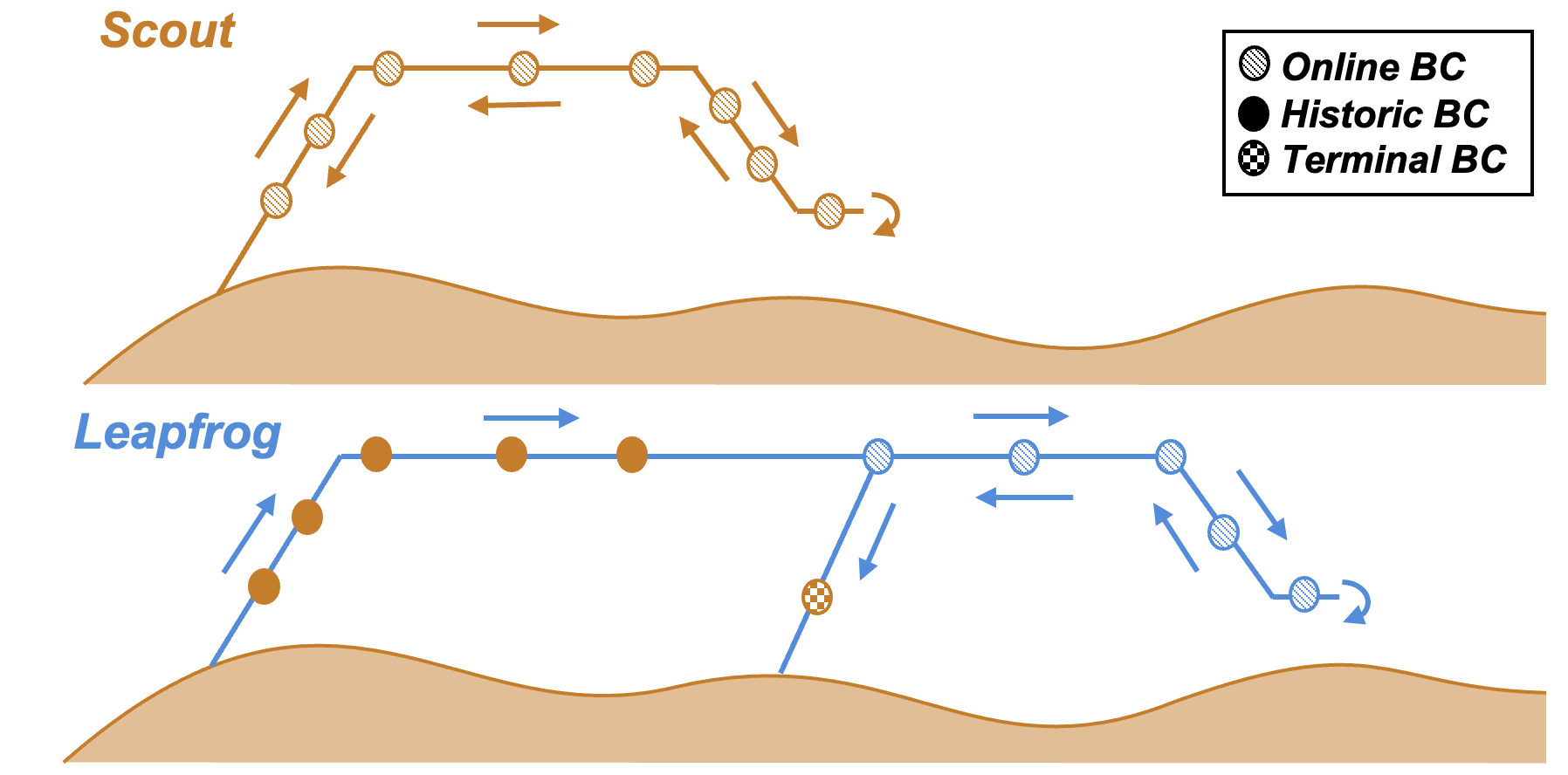}
	\caption{Mobility CONOPS and flight types}
	\label{fig:breadcrumbs}
\end{figure}

\noindent The NavCam images are processed by the \ac{ETS} module, which correlates two images to generate a lateral position displacement measurement for the nav filter.\cite{witte2019no, jenkins2022} The two \ac{NavCam} images are referred to as the $\it{current}$ image and $\it{reference}$ image. The current image is always the most recent image captured by the \ac{NavCam}, while the reference image used for correlation can be one of several types as listed in Table~\ref{tab:ets_modalities}. For recent reference images still in view of the current image, the position and velocity errors in the nav filter are highly correlated.  This position displacement measurement provides observability into velocity. Online breadcrumbs are periodically saved when one of two thresholds are crossed. The first threshold defines the maximum lateral distance scaled by the altitude from the last breadcrumb location. This threshold can also be viewed as a maximum angular or pixel offset in the camera. The second threshold type is based on a maximum scale variation defined as a maximum vertical distance over the altitude from the last breadcrumb. When an online breadcrumb is saved, the position, attitude, height \ac{AGL}, and covariance are saved in a database along with the image to allow the vehicle to retrace the path on the same flight. Following a flight, a subset of online breadcrumbs are converted into historic breadcrumbs for the next flight, which allows the vehicle to retrace segments of the previous flight. Details of this process are covered in the \nameref{design} section. The terminal breadcrumb distinction is given to images containing the landing site, where the landing site position relative to the breadcrumb image position is known to high accuracy. 
\begin{table}[htbp]
	\fontsize{10}{10}\selectfont
    \caption{ETS Measurement Modalities}
   \label{tab:ets_modalities}
        \centering 
   \begin{tabular}{|c c|} 
      \hline 
      Modality & Reference Image  \\
      \hline 
      Velocimetry & Still in NavCam FOV \\
      Online Breadcrumb & Captured earlier on current flight  \\
      Historic Breadcrumb & Captured during a previous flight \\
      Terminal Breadcrumb & Captured on current/previous flight and contains landing site  \\
      \hline
   \end{tabular}
\end{table}
Figure~\ref{fig:breadcrumbs} illustrates the primary Mobility flight types including a Scout, where the vehicle scouts a new candidate landing zone then returns to the take-off site, and a Leapfrog, where the vehicle scouts a new candidate landing zone then lands at a site that was previously scouted.
\subsection{Coordinate Frames}

Several coordinate frames are utilized by the navigation formulation. The \ac{TCI} frame is the inertial reference frame for Titan, similar to Earth-Centered Inertial.\cite{Archinal} The \ac{TCTF} is the rotating frame, with the Z-axis aligned with Titan rotation. The \ac{NED} frame is a local tangent frame where north points to the Titan spin vector (i.e. true North), down is aligned with plumb bob gravity, and east completes the right hand triad. The NED frame rotates with Titan, with the origin at the take-off position. The primary navigation frame is the \ac{TOF}, where the origin is also at the take-off position, but the frame is aligned with the $\it{estimated}$ north, east, and down directions. The TOF contains errors with respect to NED due to initial attitude errors, but by definition, does not include absolute position errors due to absolute heading error. This nuance is further discussed in the \nameref{design} section. The Dragonfly body frame uses the \ac{FRD} convention. Finally, there are instrument frames for both \acp{IMU}, the lidar, and \ac{NavCam}. 
\section{Design} 

\label{design}

\subsection{Architecture}

\begin{figure}[htb]
	\centering\includegraphics[width=6in]{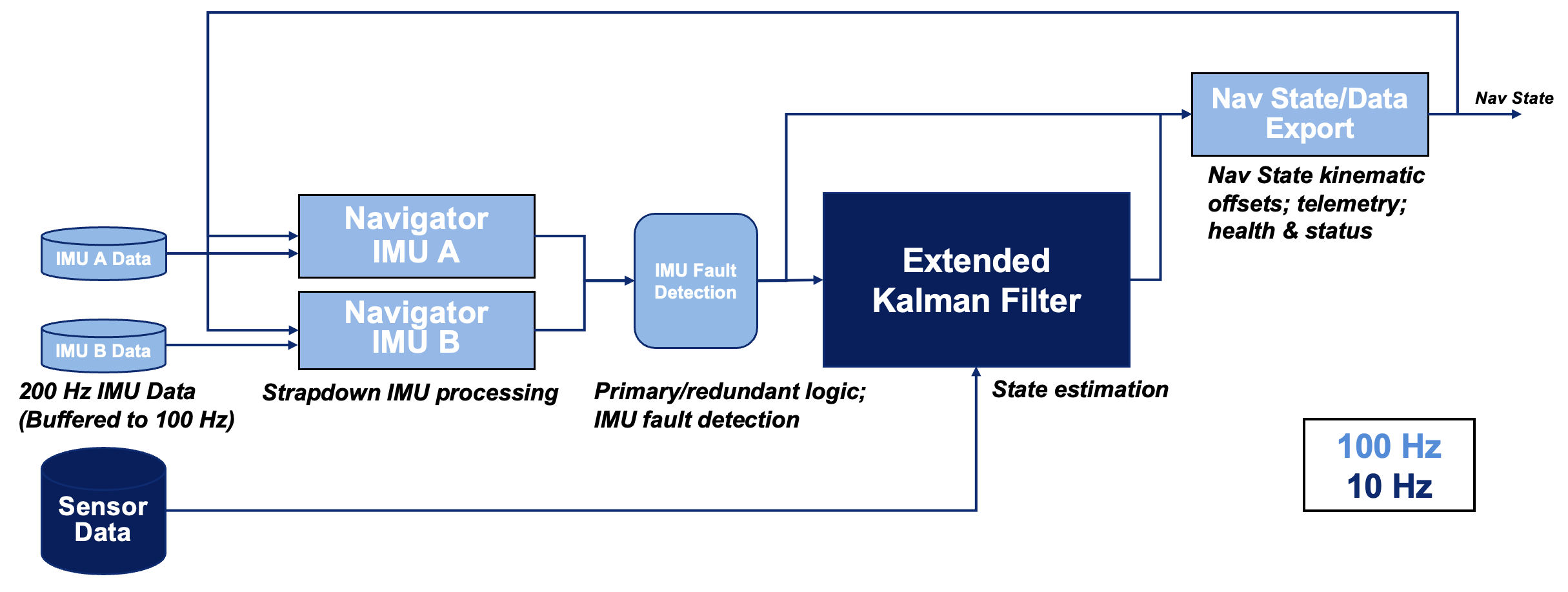}
	\caption{Navigation block diagram}
	\label{fig:nav_block_diagram}
\end{figure}

The \ac{MGNC} flight software application contains rate loops of 100 Hz, 10 Hz, and 1 Hz. As illustrated in Figure~\ref{fig:nav_block_diagram}, the navigation algorithm is split between the 100 Hz task (\ac{IMU} propagation) and 10 Hz task (Kalman filter). The 100 Hz task includes two Navigators and \ac{IMU} fault detection logic. Each Navigator propagates the vehicle pose \cite{savageAttitude, savagePosition} using data from a single \ac{IMU}. Fusing data from both \acp{IMU} into a single nav state was not pursued since the \acp{IMU} are not synchronized to a common clock, complicating the need to precisely time align inertial data.\footnote{This capability may be reevaluated in the future to address specific double IMU fault scenarios but is not covered by the baseline design.} Instead, one Navigator is considered the primary, and the other the backup. Only the state from the primary Navigator is used for navigation. The \ac{IMU} fault detection uses parity checking on integrated rate and integrated acceleration to identify a bad gyro or accelerometer, respectively. The Navigators and \ac{IMU} fault detection logic run at 100 Hz using buffered sensor data at the higher \ac{IMU} output rate of 200 Hz.  

The \ac{EKF} executes at 10 Hz and estimates the error in the inertial state from the primary Navigator using pressure, lidar, and post-processed \ac{NavCam} data. The filter design adheres to several best practices, including bias modeling, factorization, underweighting, and order invariant measurement processing.\cite{carpenter2018} Alternative architectures for redundant IMUs were considered, including running two Kalman filters in parallel, one for each Navigator, or estimating two sets of position, velocity, and orientation error states in a single filter. The architecture baselined for Dragonfly minimizes the total number of states and doesn't require a second filter. The \ac{MGNC} flight code, including the nav filter, is implemented in Simulink/Matlab, where the algorithm is exported to C code to run on the BAE RAD750 flight processor.     

\subsection{Dynamics}

There are several motivations to navigate in a local tangent plane fixed to the surface. First, there are no external inertial attitude references on the Titan surface (e.g. stars), nor do navigation maps exist for landmark identification or absolute position fixes. Further, the primary sensing modalities on Dragonfly use the Titan surface (i.e. lidar and \ac{NavCam}), simplifying the measurement models when cast in a surface fixed frame. Finally, flight command generation and path planning to surface-fixed landing sites is more intuitive in a local tangent plane than a time variant inertial frame. The dynamics for the TOF follow as:~\cite{rogers} 

\begin{equation}
\begin{aligned}
\dot{\bm{R}}_{tof}^{b} &= \bm{R}_{tof}^{b} \left[\bm{\omega}_{b|tof}^{tof}\right]_{\times} \\
\dot{\bm{r}}^{tof} &= \bm{v}^{tof} \\
\dot{\bm{v}}^{tof} &= \bm{R}_{tci}^{tof}\bm{a}^{tci} - 2\bm{\omega} \times \bm{v}^{tof} - \bm{\omega} \times \left( \bm{\omega} \times \bm{r}_{0}^{tof}\right)
\end{aligned}
\end{equation}

\noindent where $\bm{R}_{tof}^{b}$ is the \ac{TOF} to body frame \ac{DCM}, and $\bm{r}^{tof}$ and $\bm{v}^{tof}$ are the position and velocity in the \ac{TOF}, respectively, and $\left[\cdot\right]_{\times}$ is the skew symmetric operator. The dynamics are mechanized using quaternions but \acp{DCM} are presented to simplify the measurement model discussion. 

\subsection{State Space}

The \ac{EKF} state space is listed in Table~\ref{tab:state_space}. The baseline state vector contains 47 total states including position, velocity, attitude error, altitude, instrument biases, ground slope, global heading errors, and augmented position states. The position and velocity states are those of the primary \ac{IMU}. Multiplicative attitude error states are cast in the reference frame instead of the body frame used by many attitude \acp{MEKF}.~\cite{martin2010generalized, carpenter2018, markley} This error definition has the drawback of a non-intuitive interpretation of attitude error covariance, but offers several benefits for the nav filter. Attitude errors defined in the reference frame do not rotate when the vehicle changes orientation, meaning attitude errors are approximately constant between the two images used for velocimetry (modulo small errors from gyro propagation). This formulation facilitates partial state augmentation by not requiring attitude augmentation for velocimetry. A secondary benefit is consistency between attitude error and global heading error definitions, where the latter are intuitively defined in the reference frame i.e. relative to true north. Finally, computational savings can be achieved when propagating the covariance due to fewer non-zero entries in the state transition matrix~\cite{zanettiOrion}. The primary motivation for including attitude states in the filter is large initial attitude errors during \ac{EDL}. For surface flights, where the initial attitude can be precisely aligned, a formulation without attitude states was considered but this approach results in suboptimal performance and would be more challenging to properly tune. 

Separate global heading errors states are used to facilitate processing breadcrumb measurements. Prior to each flight, the initial heading is estimated from the gyro using a process known as gyrocompassing\cite{savage2000strapdown}, which serves as the only absolute heading information on Titan. However, when historic breadcrumbs are observed, absolute heading knowledge can be improved since historic breadcrumbs intrinsically contain information from gyrocompassing on a previous (independent) flight.  Navigating in the \ac{TOF}, which is aligned with the estimated \ac{NED} frame at take-off, prevents the need to update all breadcrumb positions, which are saved in an offline database, every time a historic breadcrumb measurement is processed. Instead, the filter updates the two states, $\gamma_{1}$ and $\gamma_{2}$, which represent heading error relative to true north on the previous and current flights (i.e. angles between \ac{TOF}$_{1}$ and \ac{TOF}$_{2}$ and \ac{NED} frames). Heading error accumulated within a flight is tracked in $\psi$ and is small (< 0.2$^\circ$ 3$\sigma$) relative to the initial heading error from gyrocompassing. While the two global heading error states could be combined into a single relative error state, two states are currently baselined to simplify bookkeeping between flights.

The framework uses partial state augmentation to address image processing measurement delays of $\sim$900 msec and avoid buffering and reprocessing high rate IMU and sensor data.~\cite{Bayard} To reduce the overall state size, only the position vector is augmented at the image times since attitude errors relative to gravity are small ($<$ 0.2$^{\circ}$).  Given the high quality \ac{IMU}, the growth in attitude errors between images is relatively small, and the additional error from not augmenting attitude has been shown to be acceptable. There are five position vectors in the state corresponding to the vehicle position at different images: the current image, two recent images in a \ac{FIFO} buffer for velocimetry, the most recent online breadcrumb referenced by ETS, and the most recent historic breadcrumb referenced by ETS. The motivation for maintaining two reference images and six filter states for velocimetry measurements is discussed in a later section. Instead of augmenting the position vector at every image (i.e. SLAM), the nav filter only maintains a single position vector for the most recent online and historic breadcrumb referenced by ETS. The details of how the breadcrumb states are maintained is also discussed in a later section. 

The \ac{ETS} image selection algorithm uses the 3D position estimate to select the optimal reference image for correlation.\cite{witte2019no} Vertical position errors are dominated by pressure sensor measurement error and model uncertainty relating pressure and altitude. The pressure gradient uncertainty on Titan is expected to be reduced to 2.5\% ($3\sigma$) after 1-2 Titan days of science observation. Assuming a linear model, a pressure gradient error acts as a scale factor error, manifesting as a vertical position bias that is a function of altitude (e.g. 10 m error at 400 m cruise altitude). Reducing the $\it{relative}$ vertical position error between flights is critical to ensure the \ac{ETS} algorithm selects an image that will generate a strong correlation. Estimating the atmospheric scale height error $\delta$H reduces the relative vertical position error across flights to maximize correlation scores. Ignoring modeling error would also introduce large vertical position errors during high-altitude atmospheric profiling flights (e.g. 50 m error at 2000 m AGL).

The nav filter also estimates the ground surface normal vector.  This estimate is used by the \ac{ETS} module's correlation algorithm to warp the current image into the reference image frame assuming a planar surface and pinhole camera model. Knowledge of the ground slope is also used during terminal descent to allow lidar altimetry to better estimate vertical velocity. The Mobility Flight Mode Manager (FMM), responsible for executing the flight command and contingency response within Mobility, can also use ground slope knowledge to determine when the vehicle is approaching a dune crest.\cite{Lorenz_2021}

The vast majority of states are modeled as \ac{FOGM} processes or static states with no dynamics. States with these dynamical properties are exploited in the mechanization to reduce CPU utilization on the flight hardware.\cite{carpenter2018} Additional states to model gravity errors are not currently baselined. Gravity variations near the Dragonfly \ac{EDL} site are expected to be on the order of 40 $\mu$g.\cite{DURANTE2019123} The need for explicit gravity error states will be revisited after the flight IMU is selected. 

\begin{table}[htbp]
	\fontsize{10}{10}\selectfont
    \caption{Kalman Filter State Space}
   \label{tab:state_space}
        \centering 
   \begin{tabular}{| c | c | c | c | c | c |} 
      \hline 
      Variable & Size & Description & Units & Frame & Model \\
      \hline 
     $\bm{r}$ & 3 & Primary \ac{IMU} position & m & \ac{TOF} & Navigation Equations \\ 
     $\bm{v}$ & 3 & Primary \ac{IMU} velocity & m/s & \ac{TOF} & Navigation Equations \\
     $\bm{\psi}$ & 3 & Orientation error & rad & \ac{TOF} & Navigation Equations \\
     $\rho$ & 1 & Range to current image & m & - & \ac{FOGM} \\
     $d$ & 1 & Perpendicular ground distance & m & - & \ac{FOGM} \\ 
      $\bm{b}_{aA}$ & 3 & \ac{IMU} A Accelerometer bias & m/s$^2$ & \ac{IMU} & \ac{FOGM} \\
      $\bm{b}_{aB}$ & 3 & \ac{IMU} B Accelerometer bias & m/s$^2$ & \ac{IMU} & \ac{FOGM} \\
      $\bm{b}_{gA}$ & 3 & IMU A Gyro bias & rad/s & IMU & FOGM \\
      $\bm{b}_{gB}$ & 3 & IMU B Gyro bias & rad/s & IMU & FOGM \\  
      $b_{pA}$ & 1 & Pressure altimeter A bias & Pa & - & FOGM \\
      $b_{pB}$ & 1 & Pressure altimeter B bias & Pa & - & FOGM \\
      $\bm{b}_{ETS}$ & 2 & ETS Velocimetry Bias & m & TOF or CAM & FOGM \\
      $\delta H$ & 1 & Atmospheric scale height error & m & - & FOGM \\
      $\bm{n}$ & 2 & Ground surface normal & - & TOF & FOGM  \\
      $\gamma_{2}$ & 1 & Global heading error in TOF$_{2}$ & rad & TOF & Static \\
      $\gamma_{1}$ & 1 & Global heading error in TOF$_{1}$ & rad & TOF & Static \\ 
      $\bm{r}\left(t_{c}\right)$ & 3 & IMU position at current image & m & TOF & Static \\
      $\bm{r}\left(t_{r1}\right)$ & 3 & IMU position at FIFO image \#1 & m & TOF & Static \\ 
      $\bm{r}\left(t_{r2}\right)$ & 3 & IMU position at FIFO image \#2 & m & TOF & Static \\ 
      $\bm{r}\left(t_{OBC}\right)$ & 3 & IMU position at OBC image & m & TOF & Static \\ 
      $\bm{r}\left(t_{HBC}\right)$ & 3 & IMU position at HBC image & m & TOF & Static \\
      \hline
   \end{tabular}
\end{table}

\subsection{Models}

\subsubsection{Gyrocompassing}

Initializing the nav filter prior to each flight is critical for optimal navigation performance and mission success. Accurate heading initialization ensures the vehicle flies in the desired direction and revisits historic breadcrumbs. Initialization also minimizes inertial navigation drift during the initial climb by properly accounting for correlations between attitude and \ac{IMU} errors~\cite{park}. The vehicle tilt (i.e. pitch and roll) can be initialized within several seconds using zero position and/or zero velocity pseudo-measurements, but the cross correlations between tilt and accelerometer biases can take several minutes to reach steady-state depending on the \ac{FOGM} bias time constant and magnitude of the bias relative to gravity. However, the required initialization time is dominated by estimating the vehicle heading from the gyro, a process known as gyrocompassing~\cite{savage2000strapdown}. Assuming the vehicle is static on the surface, the gyro senses Titan rotation relative to the inertial frame. The measurement model uses the following relationships of the rate in the \ac{NED} frame $\bm{\omega}^{n}$, which is a function of latitude $\phi$: 

\begin{equation} \label{eq:gyrocompass}
\bm{\omega}^{n} = \bm{R}_{TCTF}^{n} \begin{bmatrix} 0 \\ 0 \\ \Omega \end{bmatrix} = \begin{bmatrix}
\Omega\cos\phi \\
0 \\
- \Omega\sin\phi \\ 
\end{bmatrix} = \bm{R}_{b}^{n} \bm{R}_{imu}^{b}\bm{\omega}^{imu}
\end{equation}

\noindent where $\Omega$ is the magnitude of Titan rotation. Expanding Eq.~\ref{eq:gyrocompass} into estimated ($\hat{\cdot}$) and error ($\delta{\cdot}$) quantities and linearizing results illustrates that the angular rate error in the \ac{NED} frame is a combination of the attitude error and gyro error, as show in Eq. ~\ref{eq:gyrocompass_error}.

\begin{equation} \label{eq:gyrocompass_model}
\begin{aligned}
y &= 0 \\
\hat{y} &= \hat{\omega}_{east} = \hat{\bm{R}}_{b}^{n}\hat{\bm{R}}_{imu}^{b}\hat{\bm{\omega}}^{imu} + \eta \\
z &= y - \hat{y} \simeq \bm{H}\delta \bm{x} + \eta \\
\bm{H} &= \begin{bmatrix} & & \end{bmatrix}
\end{aligned}
\end{equation}

\begin{equation} \label{eq:gyrocompass_error}
\begin{aligned}
\hat{\bm{\omega}}^{n} + \delta\bm{\omega}^{n}&= \left(\bm{I} + \left[\bm{\psi}\right]_{\times}\right)\hat{\bm{R}}_{b}^{n}\hat{\bm{R}}_{imu}^{b} \left(\hat{\bm{\omega}}^{imu} + \delta \bm{\omega}^{imu}\right) \\
\delta\bm{\omega}^{n} &\approx \left[\bm{\psi}\right]_{\times}\hat{\bm{R}}_{b}^{n}\hat{\bm{\omega}}^{b} + \hat{\bm{R}}_{b}^{n}\hat{\bm{R}}_{imu}^{b}\delta \bm{\omega}^{imu} \\
\end{aligned}
\end{equation}

\noindent It is assumed in Eq.~\ref{eq:gyrocompass_error} that the error in $\hat{\bm{R}}_{imu}^{b}$ can be ignored. With rate-integrating gyros, the observed rates are computed by accumulating $\Delta\theta$ for a fixed amount of time and dividing by the accumulation time. The Kalman filter can separate heading error from gyro error as the IMU rotates in the inertial frame. With the slow rotation of Titan ($\sim$0.94$^\circ$/hr), however, it takes $\sim$16x longer to achieve the same heading accuracy as it would on Earth. The linearized measurement model for redundant IMUs follows as:

\begin{equation}
\begin{aligned}
\bm{z} = \bm{0} - \hat{\omega}_{east} = \bm{H}\delta \bm{x} + \eta = -\left[\hat{\bm{\omega}}^{n}\right]_{\times}\bm{\psi} + \hat{\bm{R}}_{b}^{n}\hat{\bm{R}}_{imu}^{b}\bm{b}_{gA} + \eta
\end{aligned}
\end{equation}

\noindent Rate measurements from both IMUs are included to reduce the heading error by $\sqrt{2}$ for a given initialization time. Dragonfly will explore latitudes near Selk Crater~\cite{barnes2021science} ($\phi$ $\sim$7.0$^\circ$) so the formulation does not account for singularities encountered near the poles.

\subsubsection{Backup IMU Biases}

Biases of the backup \ac{IMU} are estimated to improve performance if the primary \ac{IMU} fails. The primary IMU biases are only partially observable during surface initialization from the zero position and gyrocompassing measurement models because the vehicle remains at a single orientation. Full observability into the primary IMU biases is achieved in flight, from pressure and velocimetry measurements taken over multiple orientations being fused with propagated \ac{IMU} data. Data from the backup IMU is not fused into the nav state so these biases are not observable through the dynamical model. Instead, observability into the backup IMU biases is achieved by leveraging redundant information provided by two IMUs. The redundant gyro formulation proposed by Pittelkau~\cite{pittelkau} has been implemented successfully on JHU/APL missions in the past (e.g., Parker Solar Probe~\cite{kinnison2020parker}). The proposed formulation extends this model to the accelerometers and for the case where the redundant information is provided by separate IMUs that are not precisely time aligned. 

The angular rates and accelerations in the body frame (i.e., $\bm{\omega}^b$, and $\bm{a}^b$, respectively) can be expressed as:  


\begin{equation}
\bm{o}^{b} = \tilde{\bm{R}}^{\dagger}\bm{o}^{imu} 
\end{equation}

\noindent where $\bm{o}^{imu}$ is the observed angular rates or accelerations in the \ac{IMU} frame (i.e, $\bm{o}^{imu} \in \{\bm{\omega}^{imu}, \bm{a}^{imu}\}$) and $\bm{\tilde{R}}^{\dagger}$ is the pseudo-inverse of the sensor-to-body transformation, $\tilde{\bm{R}} = \begin{bmatrix} \bm{R}_{imuA}^{b} \ \bm{R}_{imuB}^{b} \end{bmatrix}^{T}.$ Given $\tilde{\bm{R}}$, there exists a null-space, $\bm{N} = \operatorname{Null}(\tilde{\bm{R}})$, such that a second observation equation for the angular rates and accelerations can be constructed as

\begin{equation}
\begin{aligned}
\bm{0} + \delta\bm{o}^{null} &= \hat{\bm{N}}^{T}\left(\hat{\bm{o}}^{imu} + \delta\bm{o}^{imu}\right). \\
\label{eq:null-space-obs}
\end{aligned}
\end{equation}

\noindent Given Eq.~\ref{eq:null-space-obs}, the null-space measurement model is presented in Eq.~\ref{eq:null-space-meas-model}, where $\bm{b}$ is a vector of the accelerometer or gyro biases, $\sigma$ is set via the vendor-specified noise characterization for the IMU and $\tau_o$ scales the observation noise by the amount of time the observations were accumulated before applying the measurement. The accumulation time $\tau_o$ is selected such that differences in integrated quantities due to relative timing jitter between the IMUs is sufficiently small.

\begin{equation}
\begin{aligned}
\bm{y} &= \hat{\bm{N}}^{T}\left(\hat{\bm{o}}^{imu} + \bm{b}\right) + \bm{\eta} \\ 
\bm{R} &= \tau_{o} \sigma^{2}\bm{I}_{6x6} 
\label{eq:null-space-meas-model}
\end{aligned}
\end{equation}

\begin{figure}[htb]
	\centering\includegraphics[width=5.5in]{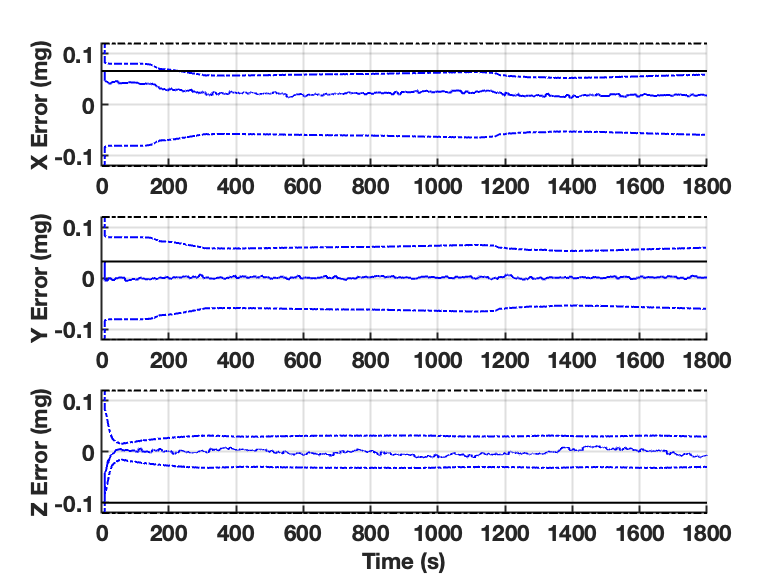}
	\caption{Observability of the backup accelerometer biases with (blue) and without (black) the null-space measurements. Solid lines represent the EKF error and dashed lines represent the EKF uncertainty.}
	\label{fig:nullspace}
\end{figure}

\noindent  Figure~\ref{fig:nullspace} illustrates observability of the backup accelerometer biases with (blue) and without (black) the null-space measurement model. The accelerometer bias time constant is set to the flight time ($\sim$ 30 minutes) and the process noise is scaled such that the steady-state markov process is aligned with the vendor specification. The solid lines represent the true accelerometer bias error and the dashed lines are the 3$\sigma$ bounds. Early observability in the Z axis is due to the \ac{IMU} vertical axis being roughly aligned with the gravity vector. Increased observability of the X and Y axes around 200 seconds and 1200 seconds come from the vehicle pitching forward to accelerate at cruise altitude and the vehicle turning around to return to the take-off location, respectively. This formulation effectively bootstraps observability of the backup \ac{IMU} biases to those of the primary \ac{IMU}. Full observability of the primary biases (via the dynamics) is transferred to the backup biases through the null-space model, ensuring the backup \ac{IMU} is primed for navigation in contingency scenarios. 

\subsubsection{Pressure} 

Dragonfly is equipped with two pressure sensors to facilitate flying constant pressure altitudes and conserving energy. The pressure sensors measure total pressure $P_{T}$: 

\begin{equation}
    P_{T} = P_{S} + P_{D}
\end{equation} \label{eq:pres}

\noindent where $P_{S}$ and $P_{D}$ are static and dynamic pressure, respectively. Static pressure can be expanded using the exponential pressure model:
\begin{equation}
    P_{S} = P_{0}\exp^{\left(\frac{h - h_{0}}{H}\right)}  
\end{equation}
\noindent where $P_{0}$ and $h_{0}$ are the reference pressure and reference height, respectively, and $h$ and $H$ are the height and scale height, respectively. The height and reference height are expanded as: 
\begin{align}
    h &= \sqrt{\left(x_{0} + x\right)^{2} + \left(y_{0} + y\right)^{2} + \left(z_{0} + z\right)^{2}} \\ 
    h_{0} &= \sqrt{x_{0}^{2} + y_{0}^{2} + z_{0}^{2}}
\end{align}
\noindent where $r_{0} = \left[x_{0} \hspace{3pt} y_{0} \hspace{3pt} z_{0}\right]$ is the position of the TOF origin with respect to Titan center in the TOF frame. Dynamic pressure is expanded as:
\begin{equation}
    P_{D} = \frac{1}{2}\rho v_{rel}^2 C_{p}\left(\alpha, \beta\right)   
\end{equation}
\noindent where $\rho$ is atmospheric density, $v_{rel}$ is the wind relative velocity magnitude, and $C_{p}$ is the coefficient of pressure, which is a function of vehicle angle of attack ($\alpha$) and sideslip ($\beta$). The linearized dynamic pressure error follows as:
\begin{equation} \label{eq:pderr}
    \delta P_{D} \approx \frac{1}{2}\left( \delta \rho \hat{v}_{rel}^2 \hat{C}_{p} + 2\hat{\rho} \hat{v}_{rel} \delta v_{rel} \hat{C}_{p} + \hat{\rho} \hat{v}_{rel}^2 \delta C_{p} + \hat{\rho} \hat{v}_{rel}^2 \hat{C}_{p}\left(\frac{\partial C_{p}}{\partial \alpha} \delta \alpha +  \frac{\partial C_{p}}{\partial \beta} \delta \beta\right)\right)
\end{equation} 
\noindent The first term in Eq.~\ref{eq:pderr} represents error in atmospheric density, the second term represents error in wind relative velocity, the third term represents error in the coefficient of pressure function, and the final term represents indexing into the coefficient of pressure function at the wrong location due to sideslip and angle of attack errors. Note the dynamic pressure error is eliminated when $C_{p}$ = 0. Since the Mobility CONOP only requires altitude knowledge from the pressure sensors, dynamic pressure errors are considered a nuisance and the desire is to minimize their magnitude by design. The pressure port, where the pressure is sensed and the $C_{p}$ function is defined, was selected to minimize $C_{p}$ and the $\frac{\partial C_{p}}{\partial \alpha}$ $\frac{\partial C_{p}}{\partial \beta}$ terms. This design alleviates the need for a dynamic pressure calibration table and additional filter states to track wind relative knowledge. Until the $C_{p}$ model has been validated via wind tunnel testing, the navigation algorithm assumes that the dynamic pressure error can be simplified and modeled using a FOGM process:
\begin{equation}
    \delta P_{D} \approx b_{p} 
\end{equation}
\noindent where $b_{p}$ soaks up modeling errors from neglecting small $C_{p}$ contributions at a subset of wind relative orientations. The assumptions and model will be revisited after collecting pressure sensor data in a wind tunnel in 2023. Finally, the measurement model for pressure altimeter A can be formed:
\begin{align}
    y       &= P_{T} = P_{S} + P_{D} = P_{0} \exp^{\left(\frac{h - h_{0}}{H}\right)} + \frac{1}{2}\rho v_{rel}^2 C_{p}\left(\alpha, \beta\right) + \eta \\
    y_{pA} &= P_{0} \exp^{\left(\frac{h - h_{0}}{H}\right)} + b_{pA} + \eta
\end{align}
\noindent Measurements from both pressure sensors are processed to reduce vertical position errors. The reference pressure $P_{0}$ is calibrated during pre-flight and can be modeled as a random walk of 0.002 mbar$/{\sqrt{s}}$, resulting in an altitude reference that can walk up to $\sim$1.5 m (3$\sigma$) over a 30 minute flight. The random walk error is considered negligible and therefore not tracked by the nav filter throughout the flight. Since the two pressure sensors share a common pressure port the measurement errors between sensors could be correlated;  modeling this correlation in the nav filter will be investigated after collecting wind tunnel data.  

\subsubsection{Lidar} The Ocellus lidar developed by \ac{GSFC} is used for hazard detection and altimetry. The nav filter uses altimetry for AGL knowledge and vertical velocity knowledge during landing, while ETS uses altimetry to convert pixel shifts into lateral displacements. Table~\ref{tab:lidar_modes} enumerates the three operational modes for the 15$^\circ$ field of view (FOV) lidar. The primary mode for surface flights is pyramid mode, where the lidar measures the 3D point along five line of sights (LOS) in the instrument frame. The +X and +Y LOSs represent +7.5$^\circ$ forward of boresight and +7.5$^\circ$ right of boresight, respectively. Altimetry mode is baselined for high altitude surface flights and initializing the nav filter on the parachute during \ac{EDL}. The high rate scanning mode is used for hazard detection operations. 
\begin{table}[htbp]
	\fontsize{10}{10}\selectfont
    \caption{Lidar Modes}
   \label{tab:lidar_modes}
        \centering 
   \begin{tabular}{| c  c  c c|} 
      \hline 
      Mode & Rate (Hz) & Line of Sight Measurements & Flight Segments \\
      \hline 
      Pyramid & 5 & Boresight, +Y, -Y, +X, -X & 15 m - 400 m \\
      Altimetry & 10 & Boresight & 400 m - 2000 m \\
      Scanning\tablefootnote{The nav filter only receives a subset of data in scanning mode.} & 120,000 & Boresight, +Y, -Y & Hazard detection scans \\
      \hline
   \end{tabular}
\end{table}
The unit vector of the 3D lidar point in the TOF frame is:
\begin{align}
    \bm{\bar{l}}^{tof} &= \bm{R}_{b}^{tof}\bm{R}_{ldr}^{b}\frac{\bm{l}^{ldr}}{\lvert \lvert \bm{l}^{ldr} \rvert \rvert} \\
    \bm{l}^{ldr} &= \begin{bmatrix} l_{x} & l_{y} & l_{z} \end{bmatrix}^{T}
\end{align}
\noindent and $\bm{R}_{ldr}^{b}$ is the instrument to body frame DCM. The measurement model assumes the ground is planar and follows as:
\begin{align} 
    y &= \frac{d}{\bm{\bar{l}}^{tof} \cdot \bm{\bar{n}}^{tof}} + \eta \label{eq:pgd} \\
    \bm{\bar{n}}^{tof} &= \begin{bmatrix} n_{n} & n_{e} & \sqrt{1 - n_{n}^2 - n_{e}^2} \end{bmatrix}^{T} \label{eq:pgd2}  \\
    \dot{d} &= - \bm{v}^{tof} \cdot \bm{\bar{n}}^{tof} + w
 \label{eq:pdg_dyn}
\end{align} 
\noindent where $\bm{n}^{tof}$ is the ground surface normal in the TOF. The ground surface normal is modeled as a random walk in distance. The FOGM power spectral density is set by assuming a max slope change (e.g. 5$^\circ$/$\sqrt{100}$ m) and the time constant is scaled to force the steady-state process to a max slope value. In altimetry mode, the ground slope is only observable from lander motion, specifically along-track knowledge from forward flight or pitching, and cross-track knowledge from banking or rolling. Observability of these states increases in pyramid and scanning modes as the lidar directly measures points off boresight, thus reducing sensitivity to lander dynamics. 

\begin{figure}[htb]
	\centering\includegraphics[width=5in]{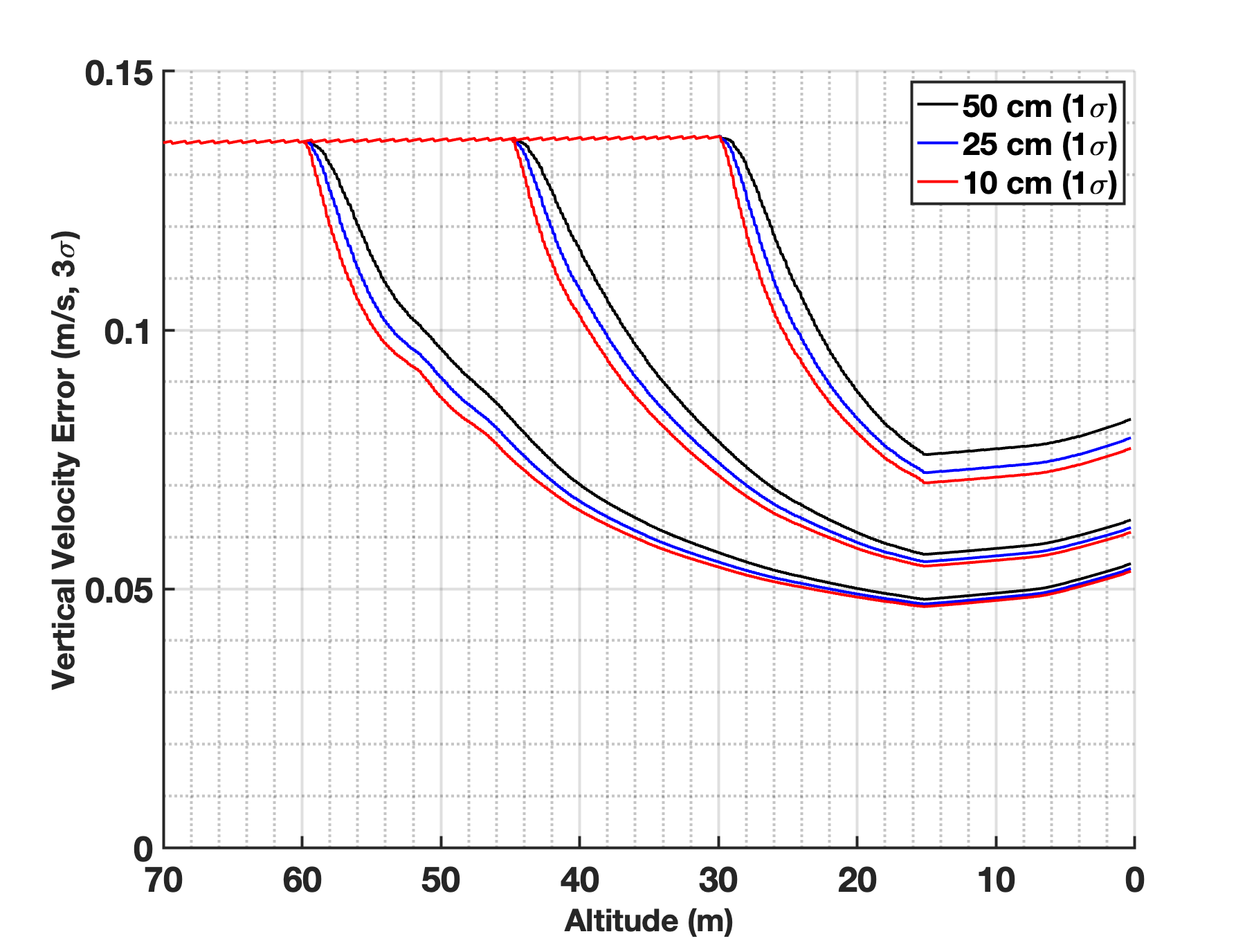}
	\caption{Lidar influence on vertical velocity knowledge during terminal descent to landing. The nine cases illustrate sensitivity to noise and altitude where the lidar is fused into the inertial nav state.}
	\label{fig:vertical_velocity}
\end{figure}

Sequential range measurements contain information about vehicle velocity but the information quality is limited by the validity of the ground model. For applications where a high fidelity ground model or \ac{DEM} is available, it is possible to robustly extract velocity knowledge from sequential ranging. A ground model isn't available for Titan so the proposed formulation accepts that the planar assumption doesn't hold over dune crests~\cite{Lorenz_2021} and prevents lidar measurements from corrupting other states by ignoring sensitivities in the measurement model (for attitude in Eq.~\ref{eq:pgd}) and dynamical model (for velocity in Eq.~\ref{eq:pdg_dyn}). However, at lower altitudes, when the planar ground model is valid, the vertical velocity partials are included in the state transition matrix. These terms allow the nav filter to extract vertical velocity knowledge from sequential ranging. Figure~\ref{fig:vertical_velocity} illustrates the vertical velocity uncertainty as a function of altitude for a nominal landing\footnote{1 m/s descent from 70 m - 20 m, gradually ramping down to 0.4 m/s by 10 m}. The steady-state vertical velocity error is driven by IMU performance and pressure sensor measurement error for the majority of the flight. Nine cases were run where the measurement noise in Eq.~\ref{eq:pgd} was varied between 10 - 50 cm 1$\sigma$ and the altitude where the lidar was fused into the inertial state (i.e. partial derivatives enabled) beginning at 30, 45, and 60 m. In all cases the lidar is not used below 15 m due to concerns of brown out and scattering, causing the vertical velocity error to increase from inertial propagation. Figure~\ref{fig:vertical_velocity} shows the lidar can reduce the vertical velocity errors by 50\% or more during the final descent to landing. Higher order ground models (e.g. quadratic) will be investigated in the future. 
\subsubsection{Velocimetry}

ETS measures the lateral displacement of the camera in the reference image frame:

\begin{equation} \label{eq:ets_fifo}
    \Delta \bm{c}^{c} = \bm{R}_{b}^{c}\bm{R}_{tof1}^{b}\left(t_{r}\right) \biggl\{\bm{R}_{tof2}^{tof1}\left(\gamma_{1}, \gamma_{2}\right)\left(\bm{r}^{tof2}\left(t_{c}\right) + \bm{R}_{b}^{tof2}\left(t_{c}\right)\bm{c}^{b}\right) - \left(\bm{r}^{tof1}\left(t_{r}\right) + \bm{R}_{b}^{tof1}\left(t_{r}\right)\bm{c}^{b}\right)\biggl\}
\end{equation}

\noindent where $\cdot$ $\left(t_{r}\right)$ and $\cdot$ $\left(t_{c}\right)$ are quantities at the reference and current image times, respectively, $\bm{c}^{b}$ is the \ac{NavCam} position in the body frame, and $\bm{R}_{b}^{c}$ is the body to \ac{NavCam} DCM. When the reference and current image are from the same flight the term representing errors between the two TOF disappears i.e. $\bm{R}_{tof2}^{tof1} = \bm{I}$. For simplicity the TOF number (e.g. $tof2$) is dropped for the remaining velocimetry discussion. Expanding Eq.~\ref{eq:ets_fifo} into estimated and error quantities and linearizing results in:

\begin{multline} \label{eq:ets_fifo_error}
    \delta \Delta \bm{c}^{c} \approx \hat{\bm{R}}_{b}^{c}\hat{\bm{R}}_{tof}^{b}\left(t_{r}\right)\biggl\{\left( \delta \bm{r}^{tof} \left(t_{c}\right) - \delta \bm{r}^{tof}\left(t_{r}\right) \right) \\ + 
    \left( \hat{\bm{R}}_{b}^{tof}\left(t_{r}\right) \left[\hat{\bm{c}}^{b}\right]_{\times}\bm{\psi} \left(t_{r}\right) -  \hat{\bm{R}}_{b}^{tof}\left(t_{c}\right) \left[\hat{\bm{c}}^{b}\right]_{\times}\bm{\psi} \left(t_{c}\right)\right)  + 
    \left[\hat{\bm{c}}^{tof}\left(t_{c}\right) - \hat{\bm{c}}^{tof}\left(t_{r}\right)\right]_{\times}\bm{\psi} \left(t_{r}\right)   \\ +
     \left( \hat{\bm{R}}_{b}^{tof}\left(t_{c}\right)\delta\bm{c}^{b} - \hat{\bm{R}}_{b}^{tof}\left(t_{r}\right)\delta{\bm{c}}^{b}  \right)                             + 
   \left[\hat{\bm{c}}^{tof}\left(t_{c}\right) - \hat{\bm{c}}^{tof}\left(t_{r}\right)\right]_{\times}\bm{\xi}\biggl\}
\end{multline}

\noindent where $\bm{c}^{tof}$ is the NavCam position in the TOF and $\bm{\xi}$ is the IMU-NavCam misalignment vector. The five terms in Eq.~\ref{eq:ets_fifo_error} represent errors in IMU lateral displacement, errors in relative orientation between the two images,  errors in the TOF to body frames, and finally errors due to IMU-NavCam misalignments. The IMU-NavCam alignments are expected to be calibrated prior to launch and assumed to remain sufficiently tight to safely ignore misalignments. For surface flights the pitch and roll errors are sufficiently small ($<$ 0.2$^\circ$) to ignore sensitivities to tilt errors, but the sensitivity to heading error is required to make the formulation consistent with receiving a relative measurement. Since attitude errors are approximately equal across images times for velocimetry, we assume $\bm{\psi} \left(t_{r}\right) \approx \bm{\psi} \left(t\right)$ in order to not require attitude augmentation at the image times. The velocimetry measurement model follows as: 

\begin{equation} \label{eq:ets_fifo_meas}
    \bm{y} = \hat{\bm{R}}_{b}^{c}\hat{\bm{R}}_{tof}^{b}\left(t_{r}\right)\biggl\{\left( \hat{\bm{c}}^{tof} \left(t_{c}\right) - \hat{\bm{c}}^{tof}\left(t_{r}\right) \right) + 
    \left[\hat{\bm{c}}^{tof}\left(t_{c}\right) - \hat{\bm{c}}^{tof}\left(t_{r}\right)\right]_{\times}\bm{\psi} \left(t\right) + \hat{\bm{b}}_{ETS} \biggl\} +  \bm{\eta}
\end{equation}

\noindent where $\bm{b}_{ETS}$ and $\bm{\eta}$ are the ETS bias and zero-mean white noise, respectively. The formulation assumes $\bm{\eta}$ represents random, uncorrelated errors in the image processing. Correlated measurement errors can be modeled using additional state augmentation~\cite{mourikis}, but for simplicity they are assumed to be small. However, errors in the image homography, dominated by imprecise knowledge of the terrain slope, can lead to correlated measurement errors, which are tracked in $\bm{b}_{ETS}$. The \ac{FOGM} time constant is selected to be commensurate with how long \ac{ETS} correlates to a single reference image, a function of altitude, lateral velocity, and \ac{NavCam} FOV. This allows the nav filter to properly compensate for correlated errors induced by nonplanar terrain. The next section discusses how the image separation distances are selected and how many velocimetry reference image states are tracked by the nav filter and ETS. 

\subsubsection{Velocimetry Trade Space}

Optimizing the image separation distances for velocimetry is necessary to minimize the accumulated position error over the flight. The optimization involves a trade between increasing distance, and thus time, between images, leading to improved velocity knowledge, at the expense of increasing the velocimetry measurement error. Homography errors due to nonplanar terrain 
are exacerbated by correlating two images over longer baselines. To characterize this effect an offline image processing analysis was run to see how far the baselines could be extended before homography errors begin to dominate the random, uncorrelated image processing errors. Next, a linear covariance analysis was run using only pressure, lidar, and velocimetry measurements to characterize the trade between image separation distance and measurement error. The third parameter in the trade is the number of reference image states to maintain in the filter for velocimetry measurements. Assuming forward flight at constant velocity $v_{0}$, the separation distance between the reference and current image grows monotonically from $v_{0} \Delta t$ to $D_{max}$, the maximum lateral distance allowed by the correlation algorithm. If only a single reference image is used the baseline will restart at $v_{0} \Delta t$ every time a new reference image is saved. However, if two reference images are maintained, the baseline only drops to $\frac{1}{2}D_{max}$. Replacing measurements at shorter baselines with longer baselines, at the expense of increasing the state space, can ultimately improve the traverse navigation performance.   

\begin{figure}[htb]
	\centering\includegraphics[width=5in]{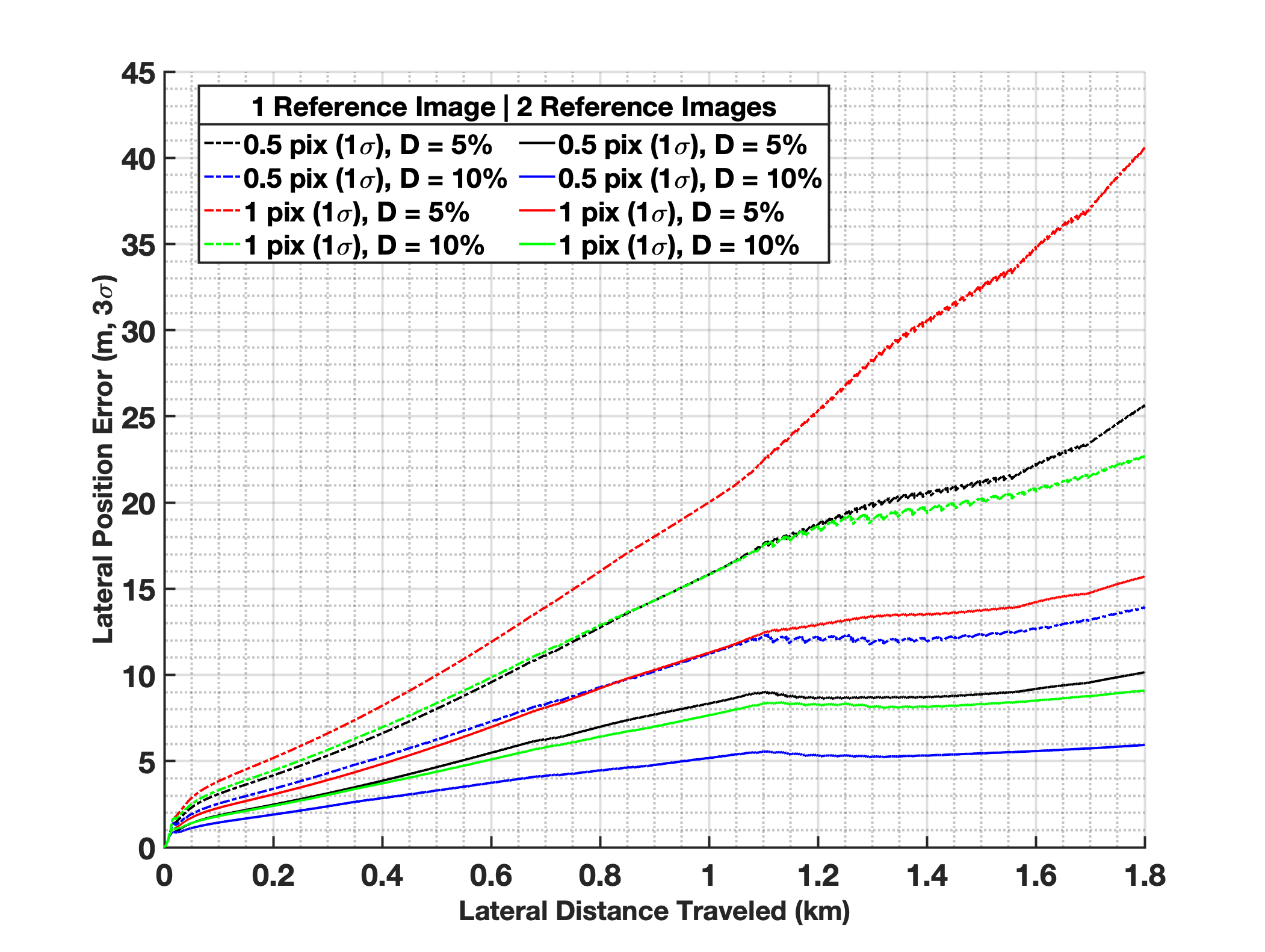}
	\caption{TOF lateral position error as a function of lateral distance flown for different configurations of image processing measurement error, image separation distance $D$, and number of reference image states maintained by the filter.}
	\label{fig:velocimetry_optimization}
\end{figure}

For the linear covariance reference trajectory, the vehicle climbs to 400 m using a flight path angle of 20$^\circ$ and traverses $\sim$1 km at $\sim$10 m/s, before descending to 100 m to scout a new candidate landing zone. Figure~\ref{fig:velocimetry_optimization} illustrates the accumulated lateral position error in the TOF as a function of distance flown for each case. The dashed lines depict how measurement noise (0.5-1 pixel 1$\sigma$) and separation distance (5 - 10\%) impact the accumulated position error using a single reference image. The solid lines depict the same configuration but when two reference images are maintained for velocimetry. The results indicate similar performance if the image baseline is increased by 2x at the expense of increasing the measurement error by 2x. However, increasing the number of reference images from one to two illustrates performance can be dramatically improved using an additional three states in the nav filter. 

\subsubsection{Breadcrumbs}

When the vehicle turns around and begins retracing the flight path, the ETS algorithm will select online breadcrumbs for correlation. When the vehicle flies out of range of the current online breadcrumb and finds the next online breadcrumb in the database, the filter must update the state and covariance for the current online breadcrumb position being tracked in the filter. Multiple approximations were explored for this step and the current implementation is discussed below.  First, the position estimate for the breadcrumb is set to the saved position in the database.  The update of the covariance presents a larger challenge to prevent collapse of the vehicle position covariance.  If the new breadcrumb position covariance is merely set to the saved position covariance in the database, and all cross covariance terms are zeroed, the filter will treat each new breadcrumb as an independent measurement.  This causes the vehicle position covariance to rapidly collapse over several breadcrumbs because it doesn't account for the high correlation between the position errors of sequential breadcrumbs.  A covariance update approach was developed to preserve as much cross correlation information between the breadcrumb and vehicle positions as possible.  This algorithm is applied separately along the north, east and down position components since a closed form solution was calculated for a single axis at a time.  

The covariance update is based on the following simplified problem.  Consider an initial covariance matrix for two variables:

\begin{equation}
    \text{P}^- = \left[ {\begin{array}{cc}
    a & 0 \\
    0 & b \\
  \end{array} } \right]
  \label{eq:blackmagic1}
\end{equation}

\noindent The goal is to define a linear measurement of the form:
\begin{align}
    y &= \left[ h_{1}\text{ }h_{2} \right] \textbf{x} + \eta \\
    \eta &\sim \mathcal{N}(0, R_{eff})
    \label{eq:blackmagic2}
\end{align}
\noindent such that after applying this measurement model to the standard Kalman filter covariance update equation, the updated covariance becomes:

\begin{equation}
    \text{P}^+ = \left[ {\begin{array}{cc}
    a_f & \frac{a_f + b_f - c }{2} \\
    \frac{a_f + b_f - c }{2} & b_f \\
  \end{array} } \right]
  \label{eq:blackmagic3}
\end{equation}

\noindent Applying the Kalman Filter update equation to Eqs. \ref{eq:blackmagic1} and \ref{eq:blackmagic2} yields:

\begin{equation}
    \text{P}^+ = \left[ {\begin{array}{cc}
    a\left(1 - \frac{ah_1^2}{z}\right) & \frac{-abh_1h_2}{z} \\
    \frac{-abh_1h_2}{z} & b\left(1 - \frac{bh_2^2}{z}\right)
  \end{array} } \right]
  \label{eq:blackmagic4}
\end{equation}

\begin{equation}
    S = ah_1^2 + bh_2^2 + R_{eff}
  \label{eq:blackmagic5}
\end{equation}

\noindent It is assumed that $a$, $b_f$ and $c$ are known.  These correspond to the vehicle position covariance for the given axis, the new breadcrumb covariance along that axis, and the cross-correlations between the breadcrumb and vehicle positions along the axis.  The problem is to solve for $b$, $h_{1}$, $h_{2}$ and $R_{eff}$.  In this formulation, $a_f$ is set to a value slightly less than $a$.  The values cannot be equal because of mathematical constraints, but setting them close accounts for the fact that switching to a new breadcrumb before receiving the measurement to it does not change the knowledge of the current vehicle position.  Under this formulation, $z$ can also be chosen arbitrarily and is set to a large parameterized number.  Setting Eqs. \ref{eq:blackmagic3} and \ref{eq:blackmagic4} equal allows the desired values to be solved:

\begin{align}
    b &= b_f + \frac{(a_f + b_f - c)^2}{4(a - a_f)} & 
    h_1 &= \sqrt{\frac{(a-a_f)z}{a^2}} \\
    h_2 &= \sqrt{\frac{(b-b_f)z}{b^2}} &
    R_{eff} &= z - ah_1^2 - bh_2^2
    \label{eq:blackmagic6}
\end{align}

\noindent This process is applied three times to the filter, once for each position direction, to update the covariance of the newly loaded online or historic breadcrumb.  When applying the measurement model for the covariance update for each axis in the actual filter, the $H$ matrix is 1x$N_{states}$ and the calculated $h_1$ and $h_2$ values are put into the indices corresponding the vehicle position and the breadcrumb position for that axis.  Although this approach is an approximate heuristic, it has shown good performance in preventing covariance collapse and maintaining the relative position errors between the vehicle position and breadcrumb positions.

After the pseudo measurement is run, the breadcrumb measurement is processed using a similar model as Eqs.~\ref{eq:ets_fifo}-\ref{eq:ets_fifo_meas}. For breadcrumbs the $\bm{b}_{ETS}$ term is not included as the bias corresponds to terrain induced errors in the images used for velocimetry, which won't necessarily align with the image geometry used for the breadcrumb correlation. Further, for historic breadcrumbs, the term representing errors between the two TOFs is expanded:

\begin{equation}
\bm{R}_{tof2}^{tof1}\left(\gamma_{1}, \gamma_{2}\right) = \bm{R}_{ned}^{tof1}\left(\gamma_{1}\right)\bm{R}_{tof2}^{ned}\left(\gamma_{2}\right) =  \bm{R}_{3}\left(\gamma_{1}\right)\bm{R}_{3}\left(-\gamma_{2}\right)
\end{equation}

\noindent where $\bm{R}_{3}$ is a rotation about the z axis. This term, and the associated sensitivities in the measurement model, is the mechanism to improve absolute heading knowledge by processing historic breadcrumb measurements.  

\subsection{Mapping Breadcrumbs Between Flights}

Since all online breadcrumb positions are defined relative to the takeoff location, breadcrumbs from previous flights that are promoted to historic breadcrumbs must be updated relative to the new takeoff location. This is done by saving the estimated landing position and covariance at the end of the flight. The curvature of Titan will also lead to slight variations in the true NED frame at each new takeoff location ($\sim$5 km resulting in changes of $\sim$0.1$^\circ$).  The rotation matrix between the true NED directions at takeoff and landing can be calculated and used to apply an additional correction.

The position of the $i^{th}$ breadcrumb in the TOF of the next flight, $\bm{r}_{bc_i}^{tof2}$ is updated by subtracting off the landing location in the original TOF, $\bm{r}^{tof1}(t_f)$ from each breadcrumb location in the original frame, $\bm{r}_{bc_i}^{tof1}$ and then rotating the point into the new NED frame at the landing site:

\begin{equation}
    \bm{r}_{bc_i}^{tof2} = \bm{R}_{tof1}^{tof2} \left( \bm{r}_{bc_i}^{tof1} - \bm{r}^{tof1}(t_f) \right)
\end{equation}

Updating the covariance of each breadcrumb requires an approximation because the cross covariance of all of the breadcrumbs to the landing position is not saved in the filter.  This approximation starts by calculating the covariance of the final online breadcrumb from the previous flight relative to vehicle position at landing and adding the approximate covariance between the final breadcrumb position and the the breadcrumb being updated, $\Delta \text{P}$.  This total covariance is then rotated into a frame corresponding to NED at the new location.  
\begin{equation}
    \text{P}_{bc}^{n_2} = \bm{R}_{n_1}^{n_2}\left(\bm{H}_{rel} \text{P}^{n_1}(t_f) \bm{H}_{rel}^T + \Delta \text{P}\right) \bm{R}_{n_1}^{n_2 T}
\end{equation}
\noindent where $\bm{P}^{n_1}(t_f)$ is the full covariance matrix at the landing time, $\bm{H}_{rel}$ is a matrix of size 3 x $\text{N}_{states},$ with the identity matrix in the indices corresponding to the current online breadcrumb position in the filter, a negative identify matrix in the vehicle position vector indices, and zeros elsewhere.  This corresponds to calculating the position of the final breadcrumb relative to the landing site.  The relative covariance between the $i^{th}$ and $j^{th}$ breadcrumb and breadcrumb saved in the filter is approximated as:
\begin{equation}
    \Delta \text{P} = \left[ {\begin{array}{ccc}
   |\text{P}_{bc_i}[1,1] - \text{P}_{bc_j}[1,1]| & 0 & 0 \\
   0 & |\text{P}_{bc_i}[2,2] - \text{P}_{bc_j}[2,2]| & 0 \\
   0 & 0 & |\text{P}_{bc_i}[3,3] - \text{P}_{bc_j}[3,3]|\\
  \end{array} } \right]
  \label{eq:DeltaP}
\end{equation}
\noindent where $\text{P}_{bc_i}[k,k]$ is element [k,k] in the 3x3 position covariance of the $i^{th}$ breadcrumb saved in the database and $bc_f$ corresponds to breadcrumb loaded into the filter at landing. This approximation produces reasonable results because the breadcrumbs are dropped in sequential fashion and the positions defined in a frame unaffected by heading uncertainties as previously described. Only breadcrumbs along the flight path of the next flight are converted into historic breadcrumbs. The heading of the online breadcrumbs is also used to down select the number of breadcrumbs. When there are both inbound and outbound breadcrumbs available, images with the same orientation that they will be viewed on the next flight are chosen.  This is done to improve image processing performance.

\section{Results} \label{results}

This section highlights sample navigation results from a high fidelity closed loop simulation. Table~\ref{tab:sim_setup} lists the relevant navigation sensor parameters used for simulation. The simulation includes full image rendering using the Rendering and Camera Emulator (RCE)~\cite{rce} and surrogate Titan terrain using a DEM of the Namib desert.\cite{schilling2019} All simulations include ETS/image processing and closed loop guidance and control.  

\begin{table}[htbp]
	\fontsize{10}{10}\selectfont
    \caption{Simulation setup for navigation performance analysis}
   \label{tab:sim_setup}
        \centering 
   \begin{tabular}{| c | c |} 
      \hline 
      Sensor Parameter & Specification (1$\sigma$)  \\
      \hline 
        Accel bias & 40 $\mu$g \\
        Accel scale factor & 120 PPM \\
        Accel velocity random walk & 0.2 mm/s/$\sqrt{hr}$ \\
        Accel acceleration random walk & 0.05 $\mu$g/$\sqrt{hr}$ \\
        Accel white noise & 0.198 mm/s  \\
        Accel misalignment & 0.03$^\circ$ per axis \\
        Gyro bias & 0.005$^\circ$/hr  \\
        Gyro scale factor & 40 PPM  \\
        Gyro angle random walk & 0.005 $^\circ$/$\sqrt{hr}$  \\
        Gyro rate random walk & 0.017$^\circ$/hr  \\
        Gyro white noise & 0.75 $\mu$rad  \\
        Gyro misalignment & 0.03$^\circ$  \\
        Lidar white noise & 2 cm  \\
        Lidar pointing & 0.05$^\circ$  \\
        Pressure white noise & 2.7 mbar  \\
        Pressure CFD model error & 6.67\%  \\
      \hline
   \end{tabular}
\end{table}

\subsection{Gyrocompassing}

\begin{figure}[htbp]
	\centering\includegraphics[width=4.5in]{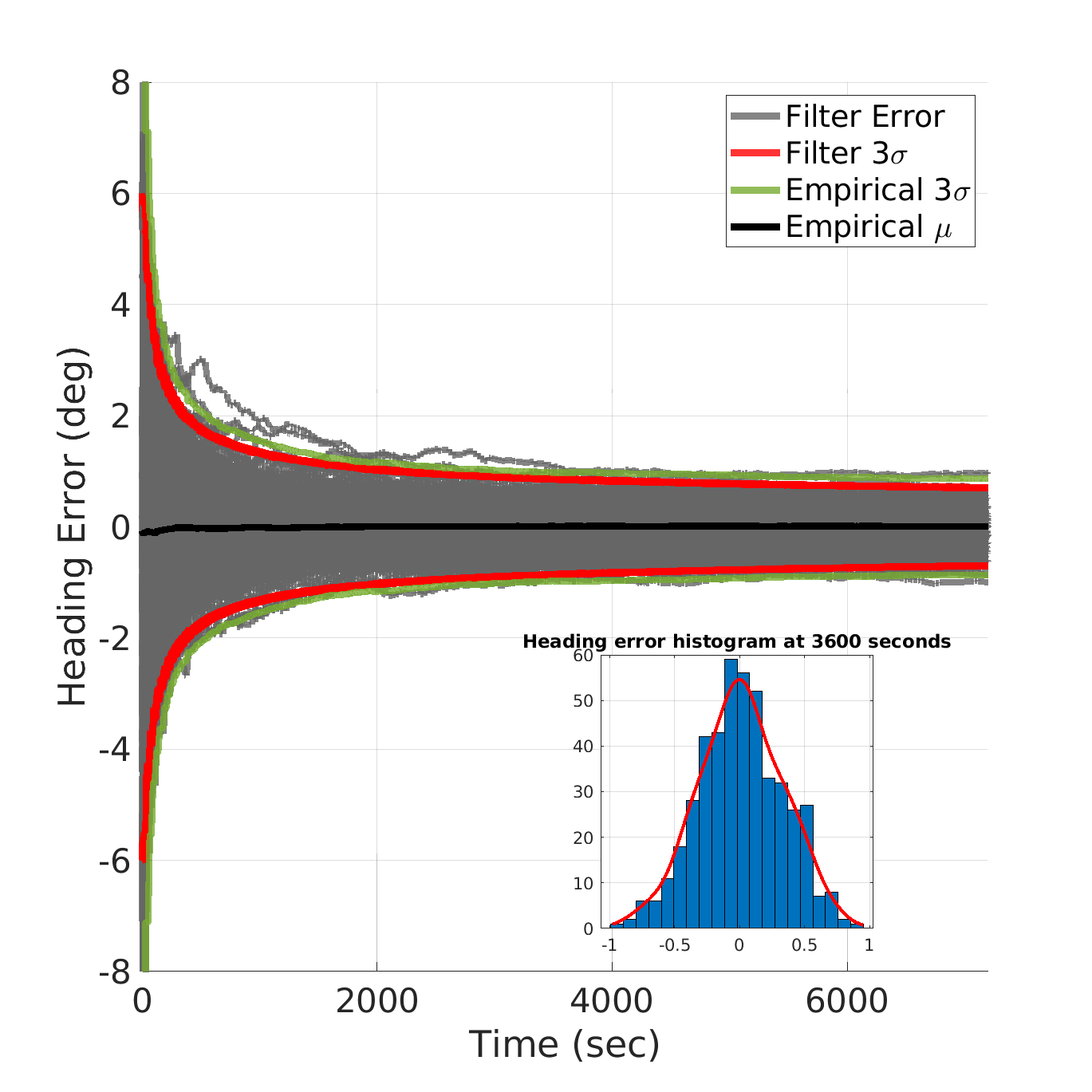}
	\caption{Heading initialization using 2 IMUs}
	\label{fig:gyrocompassing}
\end{figure}

A Monte Carlo simulation was run to characterize the gyrocompassing performance using the model described in the Design section. The initial heading error was drawn from a normal distribution with zero mean and standard deviation of 2$^\circ$, a conservative worst-case initial heading error following \ac{EDL}. 500 cases were run using parameters outlined in Table~\ref{tab:sim_setup} and the navigation filter was configured to process data from both \acp{IMU}. Figure~\ref{fig:gyrocompassing} illustrates heading error as a function of time for all cases, showing consistent errors between the empirical Monte Carlo error (green) and filter uncertainty (red). The baseline CONOP is to gyrocompass for 60 minutes, allowing the filter to squeeze heading error to less than 1$^\circ$ (3$\sigma$). 

\subsection{Leapfrog}

\begin{figure}[htbp]
	\centering\includegraphics[width=5in]{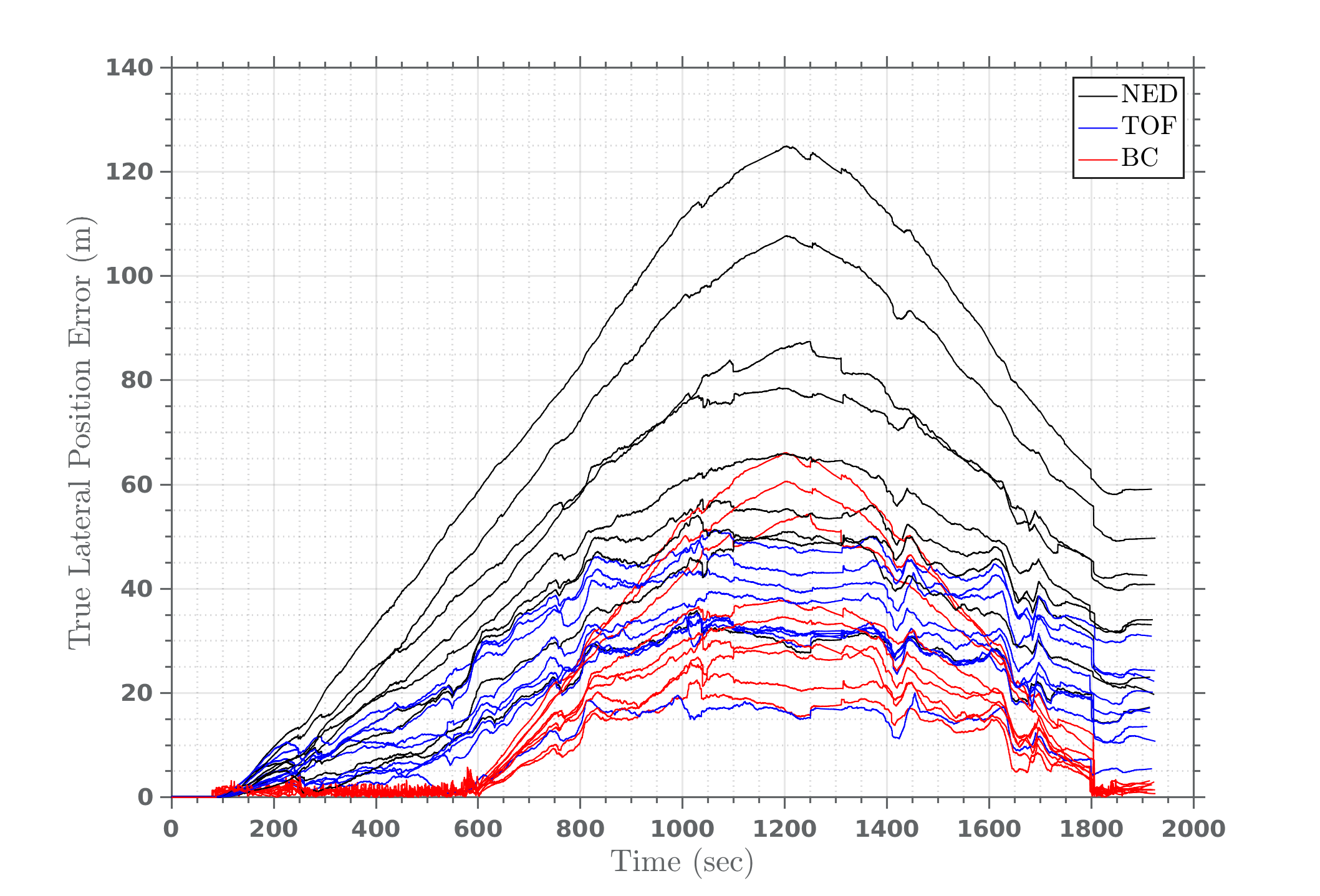}
	\caption{Lateral position errors in NED (black), TOF (blue), and relative to the historic BC in the nav filter (red).}
	\label{fig:leapfrog}
\end{figure}

A Monte Carlo of the Scout and Leapfrog were run to characterize the performance of long distance traverse navigation. The initial heading error was drawn from a normal distribution with zero mean and standard deviation of 0.4$^\circ$ for both the Scout and Leapfrog flights, representing conservative initial heading errors after gyrocompassing completes, leading to initial relative heading errors between the two flights of $\sim$1.7$^\circ$ (3$\sigma$). The root sum square (RSS) of the lateral position errors for 10 of the Leapfrog cases are illustrated in Figure~\ref{fig:leapfrog}. The colors represent lateral errors in different frames including NED (black), TOF (blue), and the breadcrumb relative error (red). When traversing down range the position errors are composed of error due to absolute heading error, which scales linearly with lateral distance, and random walk error from \ac{VIO}. Lateral errors in the NED frame represent absolute (i.e. total) position error, while errors in the TOF represent only the random walk component (by definition, the TOF does not contain position error due to global heading errors). Finally, and most important for Dragonfly, are the breadcrumb relative errors depicted in red. These errors are also in the NED frame but are relative to the historic breadcrumb in the nav filter. The breadcrumb relative errors are dominated by relative heading error between flights and a random walk between breadcrumbs. The breadcrumb relative errors remain small as the vehicle retraces the climb and cruise segments of the Scout flight but begin growing after losing historic breadcrumbs when the vehicle flies over the landing site. After scouting a new candidate landing zone the vehicle turns around and flies back towards the landing site causing the breadcrumb relative errors to decrease. Finally, when the vehicle descends and reaches the terminal breadcrumb the breadcrumb relative errors are small enough to ensure the reference image selected by ETS will contain sufficient image overlap to produce a valid correlation and displacement measurement. The breadcrumb relative errors are reduced to < 5 m after processing one or more historic breadcrumb measurements to enable precision landings.

\section{Conclusion} \label{conclusion}

The preliminary design of the Dragonfly navigation filter has been developed and demonstrated to robustly navigate the vehicle over multiple kilometer flights in a high fidelity Titan simulation. While the Dragonfly mission is unique, several contributions are applicable to the broader inertial navigation, VIO, and TRN fields including a new architecture for redundant IMUs, improving global heading and position knowledge by revisiting images after re-initializing the filter on the surface, and a robust approximation to the SLAM formulation. Several updates and optimizations are underway to improve the nav filter including porting the EKF to the UDU implementation~\cite{carpenter2018}, modeling correlated pressure sensor measurements, reducing the nav filter update rate to 1 Hz, and a higher fidelity terrain model. An alternate breadcrumb model is also being explored to better preserve cross covariance information with all the filter states. Filter updates to support image processing updates are also expected in the next phase, such as using processing multiple image patches or ingesting the displacement measurement directly in pixels.  

\section{Acknowledgment}
The authors wish to acknowledge those who improved the design and implementation through discussion, review, test, and debug: Benjamin Villac, Nishant Mehta, Jinho Kim, Steve Jenkins, Ike Witte, Sam Bibelhauser, Carolyn Sawyer, Jason Stipes, Rebecca Foust, Lev Rodovskiy, Corinne Lippe, and Lindsey Marinello. The authors also acknowledge colleagues at the University of Central Florida, Mike Kinzel and Wayne Farrell, for generating $C_{p}$ data, and Dragonfly mission architect, Ralph Lorenz.

\bibliographystyle{content/support/AAS_publication}   
\bibliography{references}   

\begin{thebibliography}{10}

\bibitem{barnes2021science}
J.~W. Barnes, E.~P. Turtle, M.~G. Trainer, R.~D. Lorenz, S.~M. MacKenzie, W.~B.
  Brinckerhoff, M.~L. Cable, C.~M. Ernst, C.~Freissinet, K.~P. Hand, {\em
  et~al.}, ``Science goals and objectives for the Dragonfly Titan rotorcraft
  relocatable lander,''  {\em The Planetary Science Journal}, Vol.~2, No.~4,
  2021, p.~130.

\bibitem{lorenzTitan}
R.~D. Lorenz, {\em Saturn's Moon Titan; Owners' Workshop Manual}.
\newblock Haynes, 2020.

\bibitem{lorenz2018dragonfly}
R.~D. Lorenz, E.~P. Turtle, J.~W. Barnes, M.~G. Trainer, D.~S. Adams, K.~E.
  Hibbard, C.~Z. Sheldon, K.~Zacny, P.~N. Peplowski, D.~J. Lawrence, {\em
  et~al.}, ``Dragonfly: A rotorcraft lander concept for scientific exploration
  at Titan,''  {\em Johns Hopkins APL Technical Digest}, Vol.~34, No.~3, 2018,
  p.~14.

\bibitem{mcgee2018guidance}
T.~G. McGee, D.~Adams, K.~Hibbard, E.~Turtle, R.~Lorenz, F.~Amzajerdian, and
  J.~Langelaan, ``Guidance, Navigation, and Control for Exploration of Titan
  with the Dragonfly Rotorcraft Lander,''  {\em 2018 AIAA Guidance, Navigation,
  and Control Conference}, 2018, p.~1330.

\bibitem{Lorenz_2021}
R.~D. Lorenz, S.~M. MacKenzie, C.~D. Neish, A.~L. Gall, E.~P. Turtle, J.~W.
  Barnes, M.~G. Trainer, A.~Werynski, J.~Hedgepeth, and E.~Karkoschka,
  ``Selection and Characteristics of the Dragonfly Landing Site near Selk
  Crater, Titan,''  {\em The Planetary Science Journal}, Vol.~2, feb 2021,
  p.~24, 10.3847/PSJ/abd08f.

\bibitem{MER}
Y.~Cheng, A.~Johnson, and L.~Matthies, ``{MER-DIMES : a planetary landing
  application of computer vision},''  2005, 2014/37440.

\bibitem{ingenuityNav}
D.~S. Bayard, ``Vision-Based Navigation for the NASA Mars Helicopter,''  {\em
  AIAA Scitech 2019 Forum}, 2019.

\bibitem{witte2019no}
I.~R. Witte, D.~L. Bekker, M.~H. Chen, T.~B. Criss, S.~N. Jenkins, N.~L. Mehta,
  C.~A. Sawyer, J.~A. Stipes, and J.~R. Thomas, ``No GPS? No Problem! Exploring
  the Dunes of Titan with Dragonfly Using Visual Odometry,''  {\em AIAA Scitech
  2019 Forum}, 2019, p.~1177.

\bibitem{schilling2019}
B.~Schilling, B.~Villac, and D.~Adams, ``Preliminary Surface Navigation
  Analysis for the Dragonfly Mission,''  {\em AAS/AIAA Astrodynamics Specialist
  Conference}, 2019.

\bibitem{smith1990estimating}
R.~Smith, M.~Self, and P.~Cheeseman, ``Estimating uncertain spatial
  relationships in robotics,''  {\em Autonomous robot vehicles}, pp.~167--193,
  Springer, 1990.

\bibitem{moutarlier1991incremental}
P.~Moutarlier and R.~Chatila, ``Incremental free-space modelling from uncertain
  data by an autonomous mobile robot,''  {\em Workshop on Geometric Reasoning
  for Perception and Action}, Springer, 1991, pp.~200--213.

\bibitem{jenkins2022}
S.~Jenkins, S.~Bibelhauser, N.~L. Mehta, J.~Thomas, and I.~R. Witte,
  ``Preliminary Flight Testing of Dragonfly's Electro-optical Terrain Sensing
  Function,''  {\em 3rd Space Imaging Workshop}, 2022.

\bibitem{Archinal}
B.~Archinal, C.~Acton, M.~A’Hearn, {\em et~al.}, ``Report of the IAU Working
  Group on Cartographic Coordinates and Rotational Elements: 2015.,''  {\em
  Celest Mech Dyn Astr}, Vol.~130, No.~22, 2018.

\bibitem{savageAttitude}
P.~Savage, ``Strapdown Inertial Navigation Integration Algorithm Design Part 1:
  Attitude Algorithms,''  {\em Guidance, Control, and Dynamics}, Vol.~21,
  No.~1, 1998, p.~130.

\bibitem{savagePosition}
P.~Savage, ``Strapdown Inertial Navigation Integration Algorithm Design Part 2:
  Velocity and Position Algorithms,''  {\em Guidance, Control, and Dynamics},
  Vol.~21, No.~1, 1998, p.~208.

\bibitem{carpenter2018}
J.~Carpenter and C.~D'Souza, ``Navigation Filter Best Practices,''  {\em
  NASA/TP–2018–219822}, 2018.

\bibitem{rogers}
R.~M. Rogers, {\em Applied Mathematics in Integrated Navigation Systems}.
\newblock American Institute of Aeronautics and Astronautics, Inc., 2007.

\bibitem{martin2010generalized}
P.~Martin and E.~Sala{\"u}n, ``Generalized multiplicative extended kalman
  filter for aided attitude and heading reference system,''  {\em AIAA
  Guidance, Navigation, and Control Conference}, 2010, p.~8300.

\bibitem{markley}
F.~L. Markley and J.~L. Crassidis, {\em Fundamentals of Spacecraft Attitude
  Determination and Control}.
\newblock Springer, 1986.

\bibitem{zanettiOrion}
R.~Zanetti {\em et~al.}, ``Absolute Navigation Performance of the Orion
  Exploration Flight Test 1,''  {\em Guidance, Control, and Dynamics}, Vol.~40,
  No.~5, 2017.

\bibitem{savage2000strapdown}
P.~G. Savage {\em et~al.}, {\em Strapdown analytics}, Vol.~2.
\newblock Strapdown Associates Maple Plain, MN, 2000.

\bibitem{Bayard}
D.~S. Bayard, ``Reduced-Order Kalman Filtering with Relative Measurements,''
  {\em Guidance, Control, and Dynamics}, Vol.~32, No.~2, 2009.

\bibitem{DURANTE2019123}
D.~Durante, D.~Hemingway, P.~Racioppa, L.~Iess, and D.~Stevenson, ``Titan's
  gravity field and interior structure after Cassini,''  {\em Icarus},
  Vol.~326, 2019, pp.~123--132, https://doi.org/10.1016/j.icarus.2019.03.003.

\bibitem{park}
H.~W. Park, J.~G. Lee, and C.~G. Park, ``Covariance Analysis of Strapdown INS
  Considering Gyrocompass Characteristics,''  {\em IEEE Transactions on
  Aerospace and Electronic Systems}, Vol.~3, No.~1, 1995.

\bibitem{pittelkau}
M.~E. Pittelkau, ``Calibration and Attitude Determination with Redundant
  Inertial Measurement Units,''  {\em Guidance, Control, and Dynamics},
  Vol.~28, No.~4, 2005.

\bibitem{kinnison2020parker}
J.~Kinnison, R.~Vaughan, P.~Hill, N.~Raouafi, Y.~Guo, and N.~Pinkine, ``Parker
  solar probe: a mission to touch the sun,''  {\em 2020 IEEE Aerospace
  Conference}, IEEE, 2020, pp.~1--14.

\bibitem{mourikis}
A.~I. Mourikis and S.~I. Roumeliotis, ``On the Treatment of Relative-Pose
  Measurements for Mobile Robot Localization,''  {\em Proceedings of the 2006
  IEEE International Conference on Robotics and Automation}, 2006.

\bibitem{rce}
C.~A. Sawyer and N.~L. Mehta, ``Rendering the Titan environment for
  Dragonfly,''  {\em 2nd RPI Space Imaging Workshop}, 2019.

\end{thebibliography}

\end{document}